\pdfoutput=1
\documentclass[10pt,twocolumn,letterpaper]{article}

\usepackage[pagenumbers]{cvpr}

\usepackage[T1]{fontenc}
\usepackage{float}
\usepackage{colortbl}
\definecolor{stagerow}{RGB}{238,245,252}
\usepackage{algorithm}
\usepackage{algpseudocode}
\usepackage{dblfloatfix}

\definecolor{cvprblue}{rgb}{0.21,0.49,0.74}
\usepackage[pagebackref,breaklinks,colorlinks,allcolors=cvprblue]{hyperref}

\title{Stage by Stage: Coordinating Vision-Language-Action Policy and Code-as-Policy for Long-Horizon Manipulation}

\author{
Yuchen Sun$^{1,2}$ \quad
Haoyu Wang$^{3}$ \quad
Hang Zhang$^{4}$ \quad
Qihang Yan$^{2}$ \quad
Qi Jia$^{2,*}$ \quad
Guangtao Zhai$^{3,*}$\\[4pt]
$^{1}$University of Science and Technology of China \quad
$^{2}$Shanghai Artificial Intelligence Laboratory\\
$^{3}$Shanghai Jiao Tong University \quad
$^{4}$Differential Robotics
}

\begin{document}
\maketitle
\begingroup
\renewcommand\thefootnote{\fnsymbol{footnote}}
\footnotetext[1]{Corresponding authors.}
\endgroup
\begin{abstract}
Hierarchical embodied agents for long-horizon manipulation increasingly leverage vision-language models (VLMs) to coordinate pretrained vision-language-action (VLA) policies and Code-as-Policy programs.
However, executor reassignment can be coupled with objective transitions, and execution context can become inconsistent across handoffs, limiting fine-grained collaboration within an unfinished objective.
We introduce \emph{STAGE}, an execution framework built on global--local coordination and selective context retention.
The VLM planner specifies stages based on observed task progress, each defined by a fixed instruction and a completion criterion. It determines whether to continue the current stage, advance to a new stage, or initiate a recovery stage.
Within each stage, VLM planner coordinates interleaved VLA and code execution at a shared observation interval measured in action steps. STAGE preserves VLA context across executor handoffs within a stage and reinitializes it from the current observation at stage transitions.
Experiments on RoboCasa365 and LIBERO-PRO demonstrate that STAGE combines the complementary strengths of VLA policies and code-as-policy programs, outperforming standalone executors and strong agentic baselines.
STAGE also achieves performance competitive with recursive self-improvement methods without relying on prior target-task exploration or the resulting memory. Further analysis highlights its potential for coordinating execution across coupled goals.
\end{abstract}

\section{Introduction}
\label{sec:intro}

Advances in visual understanding, spatial reasoning, and embodied planning are expanding the role of vision-language models (VLMs) in robotics. VLMs translate semantic goals and visual observations into actionable plans, assess task progress, and revise decisions during execution~\cite{chen2026volo,hu2026hivla}. These capabilities are particularly important for long-horizon manipulation, where success requires coordinating interdependent operations and adapting to evolving physical states. While pretrained vision-language-action (VLA) policies provide language-conditioned visuomotor skills grounded in robot experience~\cite{kim2024openvla,pi2025pi05}, long-horizon execution additionally requires organizing these skills around task dependencies and observed progress.  Hierarchical agents therefore integrate VLM planning with learned control to support skill composition, monitoring, and recovery~\cite{huang2022innermonologue,hu2026hivla}.

VLM planners access two complementary execution mechanisms: learned policies and programmatic control. VLA policies~\cite{kim2024openvla,black2024pi0} generate actions from language and visual inputs, supporting adaptive, contact-rich manipulation. Code-as-Policy programs~\cite{liang2023codeaspolicies,singh2023progprompt,huang2023voxposer} compose perception queries and analytic motion primitives to specify target poses, trajectories, and gripper operations. Hybrid systems such as HarnessVLA and VoLo expose learned policies, perception tools, and analytic primitives to a common VLM planner~\cite{zhang2026harnessvla,chen2026volo}. As physical states and control requirements evolve during execution, effective coordination between these mechanisms is essential to combining the adaptability of learned skills with the precision of geometric control.

Existing coordination interfaces suffer from two key limitations. First, restricting planner feedback to tool-call completion~\cite{zhang2026harnessvla} couples progress review and executor reassignment to coarse execution boundaries, delaying intervention when execution stalls or control requirements shift. Second, frequent VLA re-invocations disrupt instruction consistency and temporal context before an objective finishes: limited steerability degrades adherence to revised commands~\cite{chen2026steerable,park2026howlevel}, while resetting history erases context needed for temporally dependent manipulation~\cite{shi2025memoryvla}. Effective coordination therefore requires fine-grained executor transitions that preserve active objectives and valid execution context.

\begin{figure*}[t]
\centering
\makebox[\textwidth][c]{%
 \includegraphics[
   width=1.0\linewidth,
   trim=50pt 40pt 30pt 70pt,
   clip
 ]{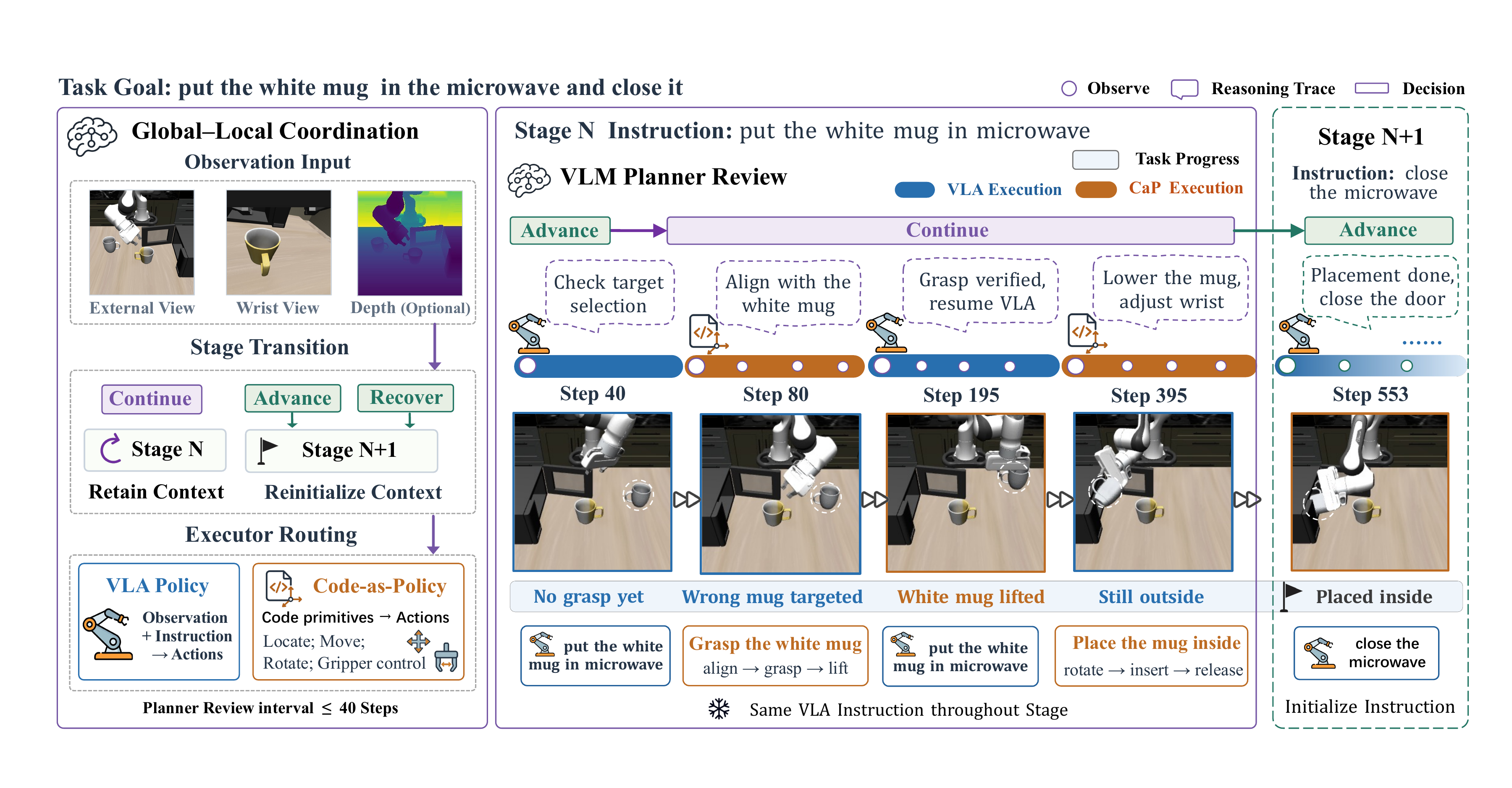}
}
\caption{\textbf{Overview of STAGE.} A persistent subgoal connects global VLM scheduling with local VLA and code execution. Observation reviews allow executor handoffs while the subgoal remains active. Selective context retention preserves compatible policy history within a stage and initializes context for each new objective.}
\label{fig:overview}
\end{figure*}

We introduce \emph{STAGE} (Fig.~\ref{fig:overview}), an inference-time execution framework that addresses these challenges through \emph{global--local coordination} and \emph{selective context retention}. The VLM planner uses task-level context and observed progress to specify stages, each defined by a fixed instruction and an observable completion criterion, and determines whether to continue, advance, or initiate recovery. Within each stage, it interleaves VLA and code execution at shared review intervals measured in action steps, allowing executor reassignment independently of stage completion. Both executors contribute toward a persistent objective through direct actions and changes to the conditions for subsequent interaction. Throughout execution, the planner maintains task progress, stage objectives, and executor outcomes to inform subsequent decisions, while VLA temporal context persists across executor handoffs within a stage. This context is synchronized with observations after code execution and reinitialized from the current observation at transitions.

We evaluate STAGE on LIBERO-PRO~\cite{zhou2025liberopro} and RoboCasa365~\cite{nasiriany2026robocasa365}. On LIBERO-PRO, STAGE achieves an average success rate of 80.0\% across Long Task and Long Swap, competitive with the reported 78.5\% of HarnessVLA with recursive self-improvement using the same VLM and VLA backbones, without prior target-task exploration or cross-episode memory~\cite{rpent2026leaderboard}. On unseen RoboCasa365 tasks, STAGE raises success from 30.0\% for the strongest evaluated agentic baseline to 53.75\%. It outperforms its VLA-only and code-only variants by 8.75\% and 37.5\%, respectively, demonstrating the benefits of combining learned manipulation skills with programmatic control. Further analyses reveal a trade-off between timely planner intervention and VLM inference cost, motivating future work on progress-adaptive review intervals. Coupled-task case studies highlight STAGE’s potential to coordinate execution across interdependent goals. These findings also motivate joint training to better align planner intent with executor capabilities and limitations for more effective collaboration.

Our contributions are:
\begin{itemize}
    \item We propose \emph{STAGE}, a framework for VLA–code collaboration through global–local coordination and selective context retention (\cref{sec:method}).
    \item Experiments across benchmarks show that improved executor collaboration enables STAGE to outperform standalone executors and strong agentic baselines, while matching or exceeding the performance of recursive self-improvement methods (\cref{sec:experiments}).
    \item Extensive analyses reveal trade-offs between timely intervention and inference cost and highlight the importance of executor collaboration in coupled tasks, motivating future work on better planner–executor alignment (\cref{sec:analysis}).
\end{itemize}

\section{Related Work}
\label{sec:related}

\paragraph{Vision-language-action models.}
A common VLA architecture combines a pretrained VLM backbone with an action expert, as exemplified by Physical Intelligence's $\pi_0$ and $\pi_{0.5}$ models with flow-matching action generation~\cite{black2024pi0,pi2025pi05}. Extending these policies to long-horizon tasks requires connecting overall instructions to intermediate objectives and actions. Demonstrations labeled only with task-level goals leave intermediate semantics implicit, while fine-grained temporal annotation is costly~\cite{shi2025hirobot,chen2026steerable}. Recent work introduces progress-based rewards to guide policy adaptation beyond demonstration imitation~\cite{bai2025evolvevla}. Complementary approaches supervise language or visual subgoals, teaching policies to predict intermediate objectives and condition action generation on them~\cite{pi2025pi05,shi2025hirobot,zhao2025cotvla}. Training on synthetic subtask, motion, and spatial commands further improves instruction following and supports hierarchical control~\cite{chen2026steerable}. These advances improve policy learning for long-horizon tasks, while coordinating pretrained VLA skills with code-based control remains complementary at execution time.

\paragraph{Programmatic robot control.}
Programmatic robot control expresses robot behavior as executable code that composes perception queries, motion primitives, and control logic. Early approaches translate language goals into robot programs~\cite{liang2023codeaspolicies,singh2023progprompt}, while subsequent work grounds execution in spatial value maps and relational keypoint constraints~\cite{huang2023voxposer,huang2024rekep}. ASPIRE refines control programs through execution feedback and consolidates validated solutions into a reusable skill library~\cite{lu2026aspire}. These approaches provide explicit geometric specification and compositional control that complement learned visuomotor skills, motivating collaboration responsive to task requirements and changing physical conditions.

\paragraph{Hierarchical embodied agents.}
Hierarchical embodied agents use high-level planners to organize low-level robot skills. Early approaches ground skill selection in affordances and adapt plans through environment feedback~\cite{ahn2022saycan,huang2022innermonologue}. Subsequent systems incorporate temporal verification~\cite{yang2025agenticrobot}, structured scene and task representations~\cite{wang2026thea,xu2026agentchord}, and transition-aware memory~\cite{xu2026baton}. Hi-VLA systematically studies planner--controller interfaces, including when control returns to the planner~\cite{hu2026hivla}. Hybrid agents broaden the planner's role to coordinating heterogeneous executors: HarnessVLA and related systems combine explicit motion with learned manipulation~\cite{zhang2026harnessvla,galbot2026gptpolicy}, while VoLo supports feedback-driven interruption and recovery across policies and action primitives~\cite{chen2026volo}. PyRUA-Lean composes robot primitives and VLA calls into programs with local checks and retries~\cite{si2026pyrua}. RoboDuo uses code to select between full-task and atomic VLA policies for execution and recovery~\cite{wang2026roboduo}. Complementary directions include joint code, video, and action modeling in AW-0~\cite{liu2026aw0} and rollout-based harness refinement with a fixed VLA executor in SERA~\cite{du2026sera}. STAGE focuses on interleaving VLA and code execution within a shared subtask objective while selectively retaining policy context across executor switches.

\section{Method}
\label{sec:method}

STAGE coordinates a pretrained VLA policy and Code-as-Policy programs to carry out long-horizon manipulation tasks. The choice of executor could change while the objective remains the same: a VLA may grasp an object, for example, before a program moves it to a computed pose. We organize execution around a \emph{stage}, which holds that objective across executor handoffs. A VLM planner reviews progress, chooses how to continue, and changes the stage when the objective is complete or requires recovery. Sec.~\ref{sec:method_setup} formalizes the execution loop. Secs.~\ref{sec:method_planning} and~\ref{sec:method_context} detail its two designs: \emph{global--local coordination}, which separates objective changes from executor reassignment, and \emph{selective context retention}, which determines what policy context persists.

\subsection{Problem Formulation}
\label{sec:method_setup}
Long-horizon manipulation requires sequential decisions over dynamic physical states. Given a task instruction $\tau$ and a horizon $T$, the agent must complete the task as judged by the native evaluator. Let $o_t$ denote the observation (images and robot state) at step $t$. A VLM planner $\mathcal{P}$ coordinates two executors: a VLA policy $\pi$ and Code-as-Policy programs combining perception with analytic motion and gripper primitives. Because both executors contribute to a shared local objective, execution requires a persistent unit bridging their calls.

We define that unit as a \emph{stage} $g=(\ell,c)$, where $\ell$ is the instruction supplied to the VLA and $c$ is an observable completion condition. The pair $(\ell,c)$ forms a persistent contract: it specifies what the current execution should accomplish and what evidence permits the planner to replace it. A stage can span multiple reviews and calls to either executor. At review $r$, the planner observes $o_{t_r}$ and returns $d_r\in\{\textsc{continue},\textsc{advance},\textsc{recover}\}$. The stage update is
\begin{equation}
g_{r+1}=\begin{cases}
g_r, & d_r=\textsc{continue},\\
\mathcal{P}(\tau,o_{t_r},g_r), & \text{otherwise}.
\end{cases}
\label{eq:stage}
\end{equation}
This transition rule unifies the initial contract instantiation and its replacement upon \textsc{advance} or \textsc{recover}. Conditioned on the resulting stage $g_{r+1}$, the planner then dispatches a sequence of executor calls for the upcoming local window. All calls between reviews share one observation interval, $t_{r+1}-t_r\le\min(W,T-t_r)$. The VLA samples chunks $a_{t:t+H}\sim\pi(\cdot\mid\ell,o_t,h_t)$ and executes a prefix, with policy context
\begin{equation}
h_t=\begin{cases}
\operatorname{Init}(o_t), & h_{t-1}=\varnothing,\ \text{VLA step},\\
h_{t-1}\oplus o_t, & h_{t-1}\neq\varnothing,
\end{cases}
\label{eq:context}
\end{equation}
where $h=\varnothing$ after each stage transition until the stage's first VLA step, and $\oplus$ denotes the backend-specific context update (\cref{sec:method_context}). Algorithm~\ref{alg:stage} instantiates this definition as a global review loop followed by bounded local execution.

\subsection{Global--Local Coordination}
\label{sec:method_planning}

Long-horizon execution couples two decisions at different scopes: the planner maintains the active local objective, while executors decide the next physical move. Executor suitability shifts dynamically—a VLA robustly handles contact-rich interactions, whereas geometric subroutines are often needed to drive an object to a computed pose. An objective typically persists until completion evidence or a concrete blocker appears. Creating a new stage at every executor switch fragments a shared subtask, whereas locking one executor to the entire stage prevents complementary collaboration. STAGE decouples these two scopes by keeping the stage contract global and executor selection local.

\paragraph{Global stage transitions.}
The global loop governs stage identity and objective progression. After initializing the contract from $\tau$ and $o_0$, the planner evaluates each review observation against the active completion condition $c$. \textsc{continue} keeps $(\ell,c)$ intact, allowing seamless switching between VLA and code execution without redefining the objective. \textsc{advance} retires the current contract and instantiates the next sub-goal once $c$ is satisfied. \textsc{recover} issues a corrective contract when obstacles, unmet prerequisites, or execution anomalies invalidate the plan, linking the new stage to its parent context.

\begin{algorithm}[t]
\caption{STAGE execution loop. Global review applies \cref{eq:stage}. Local executor routing shares an action budget and updates context by \cref{eq:context}.}
\label{alg:stage}
\small
\algtext*{EndWhile}\algtext*{EndIf}\algtext*{EndFor}
\begin{algorithmic}[1]
\Require task $\tau$, budget $T$, interval $W$; $t\gets0$, $g\gets\varnothing$
\While{$t<T$ and $\mathcal{P}$ has not finished}
  \State $d\gets\operatorname{Review}_{\mathcal{P}}(o_t,g)$ \Comment{global}
  \If{$g\neq\varnothing$ and $d=\textsc{Continue}$}
    \State \textbf{retain} $g$ and $h$
  \Else
    \State $g=(\ell,c)\gets\mathcal{P}(\tau,o_t,g)$;\; $h\gets\varnothing$ \Comment{new stage}
  \EndIf
  \State $\mathcal{P}$ selects VLA/code calls toward $c$
  \State $b\gets\min(W,T-t)$ \Comment{local}
  \While{selected calls remain and $b>0$}
    \If{the next call is VLA}
      \State \textbf{if} $h=\varnothing$ \textbf{then} $h\gets\operatorname{Init}(o_t)$
      \State $A\gets$ prefix of $a_{t:t+H}\sim\pi(\cdot\mid\ell,o_t,h)$
    \Else\ $A\gets$ next analytic primitive
    \EndIf
    \For{$a\in A$ while $b>0$}
      \State $o_{t+1}\gets\operatorname{Step}(a)$;\; \textbf{if} $h\neq\varnothing$ \textbf{then} $h\gets h\oplus o_{t+1}$
      \State $t\gets t+1$;\; $b\gets b-1$
    \EndFor
  \EndWhile
\EndWhile
\end{algorithmic}
\end{algorithm}

\paragraph{Local execution and observation interval.}
Once the global loop sets the stage, the local loop routes execution among VLA and code primitives toward $c$. Unlike methods relying on open-loop predicted durations or fixed-frequency returns~\cite{hu2026hivla}, STAGE measures local windows in physical actions. A single counter tracks all motion between reviews—encompassing VLA chunks and analytic program primitives. Once $W$ actions are spent, execution pauses to inspect a fresh observation, though a call may also return earlier. Because both executors modify the shared physical state, they operate under a unified action budget. A review serves as an observation boundary rather than a stage boundary, permitting continuation, executor handoffs, or stage transitions. We evaluate the intervention--computation trade-off of $W$ in \cref{sec:ana_window}.

\subsection{Selective Context Retention}
\label{sec:method_context}
The global--local split raises a second design question regarding which context should persist across an executor switch. While a handoff changes the executor and code may alter the physical scene, the stage objective remains valid. Selective Context Retention keeps the stage contract fixed and retains only context compatible with the resulting state. Within a stage, the instruction $\ell$ and completion condition $c$ persist across handoffs, so continued VLA calls follow the same instruction while code pursues the same criterion. The backend updates temporal context by policy type: a visual-history policy appends observations recorded during code execution or initializes from the post-code observation when code acts first, whereas an RTC policy reuses the unexecuted tail across same-stage VLA calls but discards it once code motion invalidates the prediction. A stage transition clears policy context and initializes the new stage from the current observation.

The external harness maintains a separate Planner context for progress assessment, comprising delivered observations, reasoning traces, and tool-call history. Its visual context includes the stage-start observation and up to four recent review observations. Upon a transition, this visual context resets, converting older images into text placeholders while keeping textual records across the episode. Consequently, the interface enforces three invariants: \textsc{continue} preserves $\ell$ and $c$, context crosses a handoff only if compatible with the resulting state, and a new stage initializes from fresh current-state context. Executor routing remains flexible within these constraints. Implementation details appear in the supplement, and \cref{sec:exp_ablation} evaluates these mechanisms.

\section{Experiments}
\label{sec:experiments}

\subsection{Experimental Setup}
\label{sec:exp_setup}

\begin{table*}[!t]
\centering\small
\setlength{\tabcolsep}{3pt}
\newcommand{\StageMainTableWidth}{0.94\textwidth}
\caption{\textbf{Task success (\%) on LIBERO-PRO and RoboCasa365.} CaP denotes Code-as-Policy, and RSI denotes recursive self-improvement. Bold values identify the best non-RSI agentic results, while $^{*}$ marks our reproduction method.}
\label{tab:main_results}
\makeatletter
\newcommand{\StageRowBackground}{\noalign{\begingroup
  \color{stagerow}
  \hrule width\StageMainTableWidth height\dimexpr\ht\@arstrutbox+\dp\@arstrutbox\relax
  \kern-\dimexpr\ht\@arstrutbox+\dp\@arstrutbox\relax
\endgroup}}
\makeatother
\begin{tabular*}{\StageMainTableWidth}{@{\extracolsep{\fill}}lllccc@{}}
\toprule
\multicolumn{6}{@{}c@{}}{\textbf{(a) LIBERO-PRO Long}} \\[3pt]
Method & VLM (effort) & Executors & Long Task & Long Swap & Avg SR \\
\midrule
\multicolumn{6}{@{}l@{}}{\textit{RSI methods}} \\
Zetta~\cite{ding2026zetta} & GPT-5.6-Sol (high) & $\pi_{0.5}$ + skills & 63.00 & 40.00 & 51.50 \\
HarnessVLA~\cite{rpent2026leaderboard} & GPT-6 Astra (low) & $\pi_{\text{RLinf}}$ + CaP & 85.00 & 72.00 & 78.50 \\
\midrule
\multicolumn{6}{@{}l@{}}{\textit{Standalone policies}} \\
$\pi_{0.5}$~\cite{rpent2026leaderboard} & -- & $\pi_{0.5}$ & 1.00 & 8.00 & 4.50 \\
$\pi_{\text{RLinf}}$~\cite{rpent2026leaderboard} & -- & $\pi_{\text{RLinf}}$ & 49.00 & 14.00 & 31.50 \\
\midrule
\multicolumn{6}{@{}l@{}}{\textit{Agentic methods}} \\
RoboDuo~\cite{wang2026roboduo} & Qwen3.5-2B & $\pi_{0.5}$ & 50.20 & 15.20 & 32.70 \\
AgenticRoboRL~\cite{liu2026agenticroborl} & Qwen3.5-9B & $\pi_{\text{RLinf}}$ + CaP & 41.00 & 39.00 & 40.00 \\
RPent-Hybrid$^{*}$ & GPT-6 Astra (low) & $\pi_{\text{RLinf}}$ + CaP & 56.00 & 52.00 & 54.00 \\
OpenRUA~\cite{chu2026openrua} & Fable 5.1 (high) & CaP & 70.00 & \textbf{80.00} & 75.00 \\
\StageRowBackground
\textbf{STAGE (ours)} & GPT-6 Astra (low) & $\pi_{\text{RLinf}}$ + CaP & \textbf{90.00} & 70.00 & \textbf{80.00} \\
\midrule
\multicolumn{6}{@{}c@{}}{\textbf{(b) RoboCasa365 Composite}} \\[3pt]
Method & VLM (effort) & Executors & Seen & Unseen & Avg SR \\
\midrule
\multicolumn{6}{@{}l@{}}{\textit{RSI methods}} \\
AdaRSI~\cite{wang2026adarsi} & GPT-6.1-sol & GR00T N1.5 & 49.00 & 13.80 & 31.40 \\
HarnessVLA~\cite{rpent2026leaderboard} & GPT-6 Astra (high) & RLDX-1 + CaP & 43.75 & 42.50 & 43.13 \\
\midrule
\multicolumn{6}{@{}l@{}}{\textit{Standalone policies}} \\
RLDX-1~\cite{kim2026rldx,rpent2026leaderboard} & -- & RLDX-1 & 21.30 & 5.00 & 13.15 \\
XR-1~\cite{xiaomi2026xr1leaderboard} & -- & XR-1 & 57.10 & 32.10 & 44.60 \\
\midrule
\multicolumn{6}{@{}l@{}}{\textit{Agentic methods}} \\
RPent-Hybrid$^{*}$ & GPT-6 Astra (low) & XR-1 + CaP & 13.75 & 30.00 & 21.88 \\
OpenRUA~\cite{chu2026openrua} & Claude Opus 5 & CaP & 26.30 & 28.10 & 27.20 \\
Long-WAM ~\cite{huang2026longwam} & GPT-6 Astra & Long-WAM + CaP & 38.80 & 35.00 & 36.90 \\
\StageRowBackground
\textbf{STAGE (ours)} & GPT-6 Astra (low) & RLDX-1 + CaP & 36.25 & 35.00 & 35.63 \\
\StageRowBackground
\textbf{STAGE (ours)} & GPT-6 Astra (low) & XR-1 + CaP & \textbf{76.25} & 53.75 & \textbf{65.00} \\
\bottomrule
\end{tabular*}
\end{table*}

\paragraph{Benchmarks.}

We evaluate STAGE on two widely adopted benchmarks requiring multi-step sequential reasoning. \textbf{LIBERO-PRO}~\cite{zhou2025liberopro} tests robustness to tabletop perturbations across multi-stage tasks using two 10-task suites derived from LIBERO's long-horizon benchmark: \textbf{Long Task} (altered goals) and \textbf{Long Swap} (rearranged initial object positions). \textbf{RoboCasa365}~\cite{nasiriany2026robocasa365} evaluates long-horizon household mobile manipulation across diverse kitchens on \textbf{Composite Seen} and \textbf{Composite Unseen}, chaining object manipulation, appliance operation, and navigation. We define Task SR (Success Rate) as the percentage of evaluated episodes that the benchmark's native evaluator marks as successful. Subtask SR (Subtask Success Rate) is the episode-mean fraction of native subtask conditions attained during execution. Avg SR is the unweighted mean of suite-level Task SR within each benchmark.

\paragraph{Model and execution configuration.}
We use GPT-6-Astra with low reasoning effort as the VLM planner. All VLM and VLA parameters remain frozen, and no cross-episode memory is used. For LIBERO-PRO, we use $\pi_{\text{RLinf}}$, the RLinf $\pi_{0.5}$ checkpoint fine-tuned on LIBERO-130~\cite{rlinf2025pi05checkpoint}. For RoboCasa365, we evaluate Xiaomi-Robotics-1 (XR-1)~\cite{xiaomi2026xr1} and RLDX-1~\cite{kim2026rldx}, whose task proficiency differs substantially according to their official benchmark results. STAGE uses a default observation interval of $W=40$ physical action steps. Evaluation budgets and baseline-specific execution limits are specified in Appendix~\ref{sec:supp_protocol}.

\paragraph{Baselines.}
We compare published standalone policies~\cite{rpent2026leaderboard,xiaomi2026xr1leaderboard,huang2026longwam}, RSI methods~\cite{zhang2026harnessvla,ding2026zetta,wang2026adarsi}, and embodied agents with different executors~\cite{liu2026vls,wu2026s2,wang2026roboduo,liu2026agenticroborl,chu2026openrua,si2026pyrua,wang2026auriga}. Results follow their official sources. We also include RPent-Hybrid, reproduced from the open-source HarnessVLA implementation~\cite{zhang2026harnessvla} with subgoal-conditioned VLA execution enabled. Appendix~\ref{sec:supp_protocol} provides the local reproduction protocol and published baseline configurations.

\subsection{Main Results}
\label{sec:exp_main}

Table~\ref{tab:main_results} reports task success rates on the benchmarks.

\paragraph{Highest average success among non-RSI agents.}
STAGE achieves 80\% average success on LIBERO-PRO and 65\% on RoboCasa365, outperforming the strongest non-RSI agents by 5.0 and 28.1 percentage points, respectively. With the same VLM and VLA configurations, STAGE improves over RPent-Hybrid from 54\% to 80\% on LIBERO-PRO and from 21.88\% to 65\% on RoboCasa365, highlighting the benefits of improved executor coordination. Consistent gains over standalone policies across all three evaluated VLA backbones further support its applicability across environments and executor capabilities.

\paragraph{Competitive with RSI without accumulated experience.}
On LIBERO-PRO, STAGE slightly exceeds the reported 78.5\% average success of HarnessVLA with RSI using the same VLM and VLA backbones, without prior target-task exploration or cross-episode memory. On RoboCasa365, STAGE with RLDX-1 achieves 35.63\% compared with HarnessVLA’s 43.13\%, while using lower VLM reasoning effort. These results highlight the potential of improved inference-time coordination to unlock stronger performance from non-RSI agents.
More detailed results on Subtask Success Rate and task success over executed action steps are provided in Appendix~\ref{sec:process_annotation}.

\subsection{Ablation Study}
\label{sec:exp_ablation}

We conduct controlled ablations on RoboCasa365 Composite Unseen to evaluate the contributions of STAGE's core designs: global--local executor coordination and selective context retention. Both variants use the same 80 task--seed pairs, VLA and VLM backbones, observation interval, and physical-action budgets as the full STAGE configuration. Table~\ref{tab:stage_design} reports task success, while Appx.~\ref{sec:supp_design_ablations} provides additional execution statistics and implementation details.

\begin{table}[!htbp]
\centering\small
\setlength{\tabcolsep}{3pt}
\caption{\textbf{Ablations of executor coordination and context continuity on RoboCasa365 Composite Unseen.} $\Delta$ reports Task Success Rate difference from STAGE in percentage points.}
\label{tab:stage_design}
\begin{tabular}{@{}lcc@{}}
\toprule
Configuration & Task Success Rate (\%) & $\Delta$ (pp) \\
\midrule
STAGE (ours) & \textbf{53.75} & -- \\
- Fixed executor per subtask & 36.25 & $-17.50$ \\
- Without context continuity & 35.00 & $-18.75$ \\
\bottomrule
\end{tabular}
\end{table}

\paragraph{Global--local coordination.}
This ablation asks whether executor reassignment should remain independent of stage progression. We replace STAGE's within-stage executor reassignment with a fixed executor for each subtask while keeping the planner, XR-1 backbone, review interval, context policy, and action budgets unchanged. Task SR decreases from 53.75\% to 36.25\%, a drop of 17.50 percentage points. In the matched STAGE cohort, the execution traces contain 480 within-stage VLA--Code switches across 288 mixed-executor stages, whereas the fixed-executor control has no within-stage switches and invokes Code in only 2/80 episodes. The result supports separating global objective transitions from local executor reassignment, allowing either executor to contribute to the ongoing stage objective.

\paragraph{Selective context retention.}
This ablation asks whether repeated executor calls within an unfinished stage can remain effective when each VLA call receives a newly written instruction and a fresh cross-call observation context. We remove context continuity at VLA-call boundaries while preserving the stage identifier, completion criterion, executor switching, review interval, and action budgets. Task SR decreases from 53.75\% to 35\%, a drop of 18.75 percentage points. The ablation nevertheless invokes Code more frequently: it produces 11.44 Code segments per episode and 11.73 per Code-active episode, compared with 6.64 and 8.05 for STAGE, respectively. Thus, in this matched evaluation, additional Code execution does not recover the performance lost when context continuity is removed.

\section{Analysis}
\label{sec:analysis}

\subsection{Executor Complementarity}
\label{sec:ana_executor}
\label{sec:ana_executor_patterns}

\textbf{Joint access to VLA and code improves task completion over either executor alone.}
We first examine the benefit of combining both executors within STAGE. As shown in Tab.~\ref{tab:executors}, joint execution achieves 53.75\% Task SR and 78.50\% Subtask SR, outperforming VLA-only execution by 8.75\% and 10.92\%, respectively. Code-only execution reaches only 16.25\% Task SR and 42.08\% Subtask SR. Its lower step count stems from frequent failures and early termination rather than execution efficiency. These results highlight the benefits of executor collaboration and demonstrate that STAGE effectively integrates their strengths.

\begin{table}[!htbp]
\centering\small
\setlength{\tabcolsep}{3pt}
\caption{\textbf{Executor restriction within STAGE.}}
\label{tab:executors}
\begin{tabular}{@{}lccc@{}}
\toprule
Available executors & \shortstack{Task SR\\(\%)} & \shortstack{Subtask SR\\(\%)} & Steps \\
\midrule
Code-as-Policy & 16.25 & 42.08 & 1,216 \\
VLA & 45.00 & 67.58 & 1,879 \\
VLA + Code-as-Policy & \textbf{53.75} & \textbf{78.50} & 1,929 \\
\bottomrule
\end{tabular}
\end{table}

\begin{figure}[!htbp]
\centering
\IfFileExists{figures/executor_progress.pdf}{
  \includegraphics[
    width=\dimexpr\columnwidth-2pt\relax,
    trim=1pt 12pt 1pt 24pt,
    clip
  ]{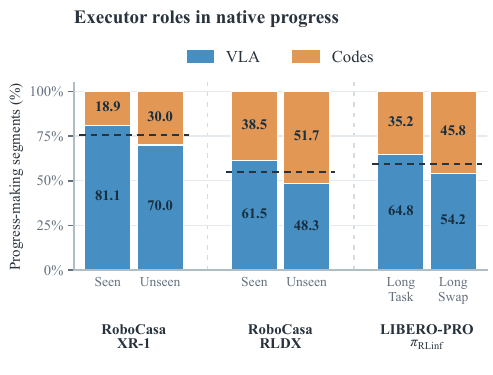}
}{
  \parbox[c][3.4cm][c]{0.94\columnwidth}{\centering\small
    \textbf{Figure asset: Executor roles in native progress}\par\medskip
    Add \texttt{figures/\detokenize{executor_progress.pdf}} to display the figure.
  }
}
\caption{\textbf{Executor contributions to task progress.}
Bars show VLA and code shares in progress-making segments. Dashed lines mark mean VLA share across suites per benchmark--policy pair.}
\label{fig:executor_progress}
\end{figure}

\textbf{Lower standalone VLA proficiency leads to a greater relative contribution from code to task progress.}
We next evaluate how executor contributions shift with VLA capability. On RoboCasa365, standalone SR are 21.30\% (Seen) and 5.00\% (Unseen) for RLDX, compared with 57.10\% and 32.10\% for XR-1 (Tab.~\ref{tab:main_results}). Figure~\ref{fig:executor_progress} breaks down progress-making segments consecutive calls to a single executor yielding net native-condition gains—across both successful and failed trajectories. Under the same STAGE protocol and VLM planner, code accounts for 38.5\% and 51.7\% of these segments with RLDX, versus 18.9\% and 30.0\% with XR-1.
STAGE thus tends to rely more on programmatic control to compensate for limited VLA proficiency.

\textbf{Code contributions scale with task requirements, particularly when spatially specified operations provide concrete targets.}
On LIBERO-PRO, code accounts for 35.2\% and 45.8\% of progress-making segments on Long Task and Long Swap, respectively. Its native-condition gains are concentrated in placing objects on surfaces or into containers, alongside closure operations. Because these operations feature explicit physical targets, they are directly amenable to programmatic control, explaining code's substantial contribution in these benchmarks.

\subsection{Progress Beyond Stage Boundaries}
\label{sec:ana_coordination}
\label{sec:ana_stage_switching}

\begin{figure}[!htbp]
\centering
\includegraphics[width=\columnwidth]{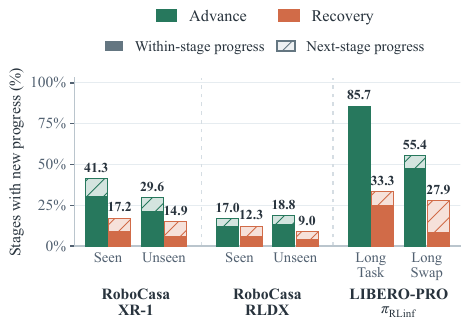}
\caption{\textbf{Progress during and after stage execution.}}
\label{fig:stage_progress_resumption}
\end{figure}

Figure~\ref{fig:stage_progress_resumption} compares progress within and immediately after Advance and Recovery stages. Solid portions represent stages with first-time native-condition attainment, while hatched portions represent stages with no such attainment whose immediate successor makes new progress.
\textbf{Progress associated with recovery often emerges in the subsequent stage.} On LIBERO-PRO Long Swap, including the immediate successor raises the fraction of Recovery stages associated with new progress from 8.2\% to 27.9\%, compared with 47.7\% to 55.4\% for Advance stages. Evaluating recovery solely by within-stage attainment therefore overlooks progress observed in subsequent execution. These observations motivate future work on understanding and accounting for a stage’s downstream effects when coordinating execution.

\begin{figure}
    \centering
    \includegraphics[width=0.89\linewidth]{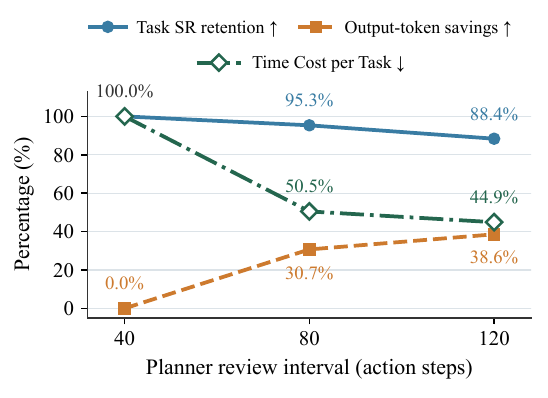}
    \caption{\textbf{Trade-off between task success and inference cost on RoboCasa365 Composite Unseen.}}
    \label{fig:interval}
\end{figure}

\begin{figure*}[!t]
\centering
\setlength{\fboxrule}{1pt}
\setlength{\fboxsep}{0pt}
  \includegraphics[
    trim=0pt 53pt 0pt 1pt,
    clip,
    width=\dimexpr0.95\textwidth-2\fboxrule-2\fboxsep\relax
  ]{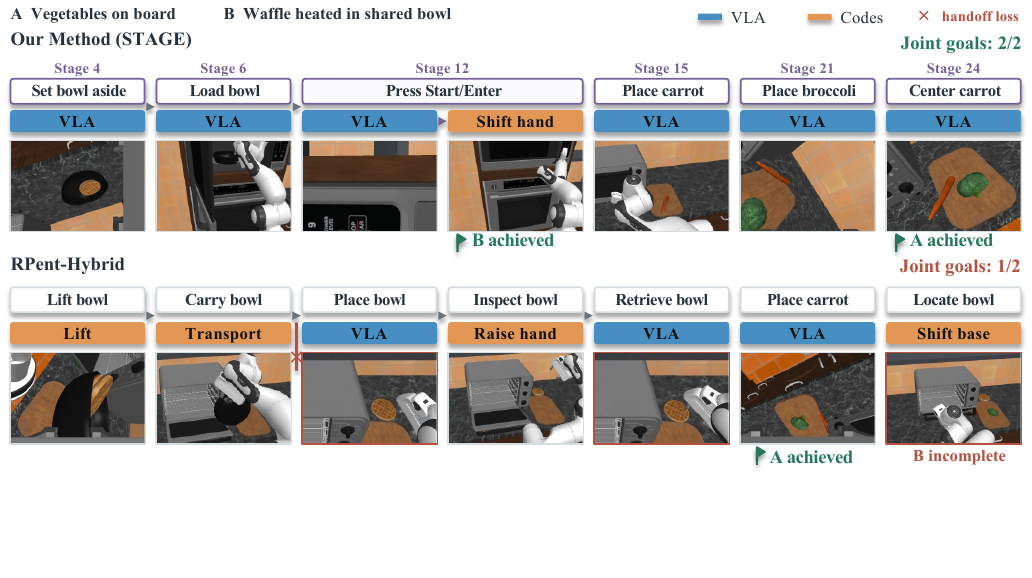}
\caption{\textbf{Task completion across coupled goals.}
}
\label{fig:coupled_goals}
\end{figure*}

\subsection{Generalization Across VLMs and VLAs}

\begin{table}[!htbp]
\centering\small
\setlength{\tabcolsep}{3pt}
\caption{\textbf{Generalization across planners and VLA backbones on RoboCasa365 Composite Unseen.} Task SR (\%) over 80 valid episodes per method.}
\label{tab:backbone_generalization}
\begin{tabular}{@{}llccc@{}}
\toprule
Variation & VLM (effort) & VLA & \shortstack[l]{RPent\\Hybrid} & \shortstack[l]{STAGE\\(ours)} \\
\midrule
Default & \shortstack[l]{GPT-6 Astra (low)} & XR-1 & 30.00 & \textbf{53.75} \\
VLM & \shortstack[l]{GPT-5.6 Sol (medium)} & XR-1 & 22.50 & \textbf{27.50} \\
VLA & \shortstack[l]{GPT-6 Astra (low)} & RLDX & 8.75 & \textbf{35.00} \\
\bottomrule
\end{tabular}
\end{table}

\textbf{STAGE consistently improves performance across the evaluated planner and VLA configurations.} As shown in Tab.~\ref{tab:backbone_generalization}, STAGE outperforms RPent-Hybrid by 23.75\% with GPT-6-Astra and XR-1, and by 26.25\% when replacing XR-1 with RLDX-1. Replacing the planner with GPT-5.6 Sol retains a 5.00\% improvement. These results support the applicability of STAGE across different planners and executors, although the varying gains indicate that coordination performance remains dependent on component capabilities.

\subsection{Success-Cost Trade-off in Review Intervals}
\label{sec:ana_window}

We vary the planner review interval from 40 to 80 and 120 physical action steps, keeping the remaining settings fixed on RoboCasa365 Composite Unseen. Figure~\ref{fig:interval} reports Task SR retention, output-token savings, and relative time cost. Task SR and time cost are normalized to their 40-step values, while token savings measure the relative reduction in mean planner output tokens. Time cost includes service waiting and tool execution. Detailed per-task metrics are provided in Appendix~\ref{sec:appendix_interval}.

\textbf{Moderately longer review intervals reduce token and time costs while preserving most of the baseline task success.} At 80 steps, STAGE retains 95.3\% of the baseline Task SR while saving 30.7\% of output tokens and reducing time cost to 50.5\% of the baseline. Increasing the interval to 120 steps raises token savings to 38.6\% and further lowers relative time cost to 44.9\%, but Task SR retention drops to 88.4\%. Thus, extending the interval beyond 80 steps yields smaller additional cost reductions at a larger loss in task success. This trade-off motivates future work on adaptive review intervals that account for execution complexity and task progress.

\subsection{Case Studies on Task Composition}
\label{sec:ana_coupled}

To examine coordination across coupled goals, we focus on a shared-object case in which a bowl both blocks vegetable placement and is needed for waffle reheating. The planner must resolve this dependency while preserving the condition required by the remaining goal, exposing whether progress on one goal survives subsequent execution. We use this high-coupling diagnostic case from RoboCasa365 Composite
(Fig.~\ref{fig:coupled_goals}) to trace the shared-object condition across the two goals. Additional coupled or single task case studies are provided in Appendix~\ref{sec:supp_coupled_protocol}.

\textbf{STAGE coordinates executor switches around shared-object dependencies.} STAGE first moves the bowl away from the board, loads it into the microwave, and completes reheating. It then resumes vegetable placement while the reheating condition remains satisfied, achieving joint completion. RPent-Hybrid grasps the bowl but loses it while transporting it to the microwave, before the waffle can be reheated. It later places the vegetables, but the lost bowl prevents reheating from being completed, leaving only one goal satisfied. The methods diverge at executor switches. STAGE establishes the condition required by the shared-object goal before returning to the remaining goal, then carries that condition through the stage transition. RPent-Hybrid makes local transport progress without protecting the reheating condition. When vegetable placement later succeeds, the missing condition cannot be recovered. This case reveals the source of STAGE's advantage on coupled goals: executor transitions preserve the conditions required by the remaining goal, so subsequent stages inherit a state compatible with joint completion.

\section{Conclusion}
\label{sec:conclusion}

We presented STAGE, an execution framework combining online stage progression, fine-grained VLA--code coordination, and selective context retention. A fixed instruction and completion criterion define each stage, while periodic reviews determine executor allocation and objective transitions. VLA context persists within a stage and is reinitialized for task progression or recovery. Experiments on RoboCasa365 and LIBERO-PRO demonstrate improved task success and native subtask-condition attainment with the Astra planner. Observation interval experiments quantify the computation--execution trade-off, and coupled-task cases demonstrate coordination across interdependent objectives. These findings establish stage-based execution as an effective coordination framework for frozen policies and programmatic control in evaluated long-horizon settings.

{
    \small
    \bibliographystyle{ieeenat_fullname}
    \bibliography{main}
}

\appendix
\clearpage
\twocolumn[
  \begin{center}
    \Large\textbf{Appendix}
    \vspace{1.0em}
  \end{center}
]
\nocite{vls2026code}
\section{Evaluation Protocol and Baselines}
\label{sec:supp_protocol}

\paragraph{Local comparison protocol.}
RPent-Hybrid is reproduced from the open-source HarnessVLA implementation~\cite{zhang2026harnessvla}. Subgoal-conditioned VLA instructions are enabled for the local RoboCasa comparison to align the language interface with LIBERO. Both local RPent-Hybrid and STAGE reuse the RPent Code-as-Policy codebase, program-to-primitive interface, primitive library, camera views, and explicit termination tool (which reports either task completion or a stuck state). The planner prompt supplies the active subgoal and completion criterion, and the Code-as-Policy executor receives the resulting local instruction through this shared interface. The local baseline retains its original executor-choice guidance, invocation protocol, and context-management rules, while STAGE wraps the same calls with stage state and transitions.

\paragraph{Benchmark execution settings.}
Following hierarchical agentic evaluation practice~\cite{zhang2026harnessvla}, RoboCasa365 local evaluations are conducted on the target kitchens with five seeds per task and a per-task physical-action budget scaled by a factor of 1.6 to accommodate multi-stage recovery and interleaved programmatic motion. Composite Seen and Composite Unseen denote the seen and unseen task splits evaluated in these kitchens. LIBERO-PRO evaluations cover ten Long Task and ten Long Swap tasks with five seeds per task and a 2,000-action budget. The LIBERO policy uses Real-Time Chunking (RTC)~\cite{black2025rtc} with a prediction horizon of ten actions and a five-action execution prefix.

\paragraph{Evaluation and accounting.}
Each task--seed pair contributes one selected outcome, including valid failures. Native task success is determined by the benchmark evaluator independently of planner-declared completion. Mean action steps, planner decisions, output tokens, and wall-clock measurements are averaged over both successful and failed episodes.

\paragraph{RoboCasa Seen failure profile.}
In Table~\ref{tab:main_results}, RPent-Hybrid achieves 13.75\% on RoboCasa365 Composite Seen compared with 30.00\% on Composite Unseen. Inspecting the 69 unsuccessful Seen rollouts shows that none exhausts the physical-action budget: 64/69 end with the planner explicitly declaring a \texttt{stuck} state after coarse subgoal attempts stall (with only 4/69 false \texttt{success} claims and 1/69 exiting without a \texttt{finish} call). On average, RPent-Hybrid executes 1,280 physical actions, 8.79 subgoal calls, 14.86 planner decisions, 2.30 Code calls, and 6.49 VLA calls per episode (42.9\% final subtask-condition attainment), whereas STAGE averages 1,675 actions, 47.58 subgoal calls, 57.30 planner decisions, 14.44 Code calls, and 33.14 VLA calls (82.5\% attainment). Thus, RPent-Hybrid's lower Seen score relative to Unseen stems primarily from coarse subgoal execution stalling into unrecovered \texttt{stuck} states rather than action-budget exhaustion.

\subsection{Published Baseline Configurations}
\label{sec:supp_baselines}

Table~\ref{tab:main_results} in the main paper reports each baseline's planner, executor, and task success rate. Below we summarize the evaluation protocols and source configurations of the published baselines, which are taken directly from their cited sources rather than rerun in our local harness.

\paragraph{Published LIBERO-PRO configurations.}
Each LIBERO-PRO row follows the planner, policy, and training or exploration setup reported by its source: Zetta~\cite{ding2026zetta} evaluates 20 held-out seeds per task; HarnessVLA~\cite{rpent2026leaderboard} uses the same GPT-6 Astra (low) planner, $\pi_{\text{RLinf}}$ policy, and analytic primitive library as STAGE over 100 episodes per Long suite, while additionally retaining frozen target-suite exploration memory; OpenRUA~\cite{chu2026openrua} evaluates 100 episodes per suite; AgenticRoboRL~\cite{liu2026agenticroborl} uses its planner and policy trained with SFT and RL (distinct from its untrained 27B reference); and RoboDuo~\cite{wang2026roboduo} uses its five-suite $\pi_{0.5}$ configuration with fine-tuned coding and judge models over 500 trials per suite. Standalone $\pi_{0.5}$ and $\pi_{\text{RLinf}}$ rows follow the RPent leaderboard~\cite{rpent2026leaderboard} without a VLM planner.

\paragraph{Published RoboCasa365 configurations.}
Each RoboCasa365 row likewise retains its source-reported setup: AdaRSI~\cite{wang2026adarsi} evaluates 50 scenes per task; HarnessVLA~\cite{rpent2026leaderboard} uses GPT-6 Astra (high) with RLDX-1, analytic primitives, and frozen exploration-derived memory over 80 episodes per Composite split; OpenRUA~\cite{chu2026openrua} evaluates ten seeds per task; and Long-WAM~\cite{huang2026longwam} uses its released pretraining-kitchen configuration. Standalone RLDX-1~\cite{kim2026rldx} follows the RPent leaderboard~\cite{rpent2026leaderboard}, while standalone XR-1 follows its official RoboCasa365 leaderboard submission~\cite{xiaomi2026xr1leaderboard}.

\paragraph{Local matching.}
Published rows retain their source-reported task splits, environment splits, experience conditions, and trial counts as system-level reference points, whereas controlled coordination comparisons (STAGE, RPent-Hybrid, single-executor restrictions, and design ablations) use matched local scenes, seeds, planners, action budgets, and evaluation schedules.

\section{Subtask Progress and Recovery Analysis}
\label{sec:process_annotation}

Subtask-condition attainment measures partial task progress beyond binary task success. We compute this metric via offline replay of saved rollout trajectories without issuing any additional policy or planner calls, evaluating a fixed set of native goal conditions for each task (native goal conditions for RoboCasa365 and BDDL goal conditions for LIBERO-PRO). For each rollout episode $i$, let $c_{ij}(t)\in\{0,1\}$ indicate whether goal condition $j$ is satisfied at action step $t$, and let $K_i$ denote the total number of evaluated conditions for that task. Its process attainment is
\begin{equation}
    P_i=\frac{1}{K_i}\sum_{j=1}^{K_i}\max_{0\leq t\leq T_i}c_{ij}(t),
\end{equation}
where $T_i$ is the step of first native task success or the episode end for a failure. We average $P_i$ equally across episodes. Each condition is counted individually if satisfied at any evaluated step along the trajectory (including $t=0$), irrespective of whether it remains satisfied later; $P_i$ thus measures cumulative subtask progress, complementing native Task SR (which requires all goal conditions to hold jointly) and endpoint condition attainment. Across all four suites, STAGE also improves endpoint condition attainment over RPent-Hybrid (90.0\% vs.\ 67.0\% on Long Task, 75.0\% vs.\ 63.3\% on Long Swap, 82.5\% vs.\ 42.9\% on Composite Seen, and 68.0\% vs.\ 46.8\% on Composite Unseen), indicating that intermediate gains are largely preserved rather than transiently lost.

\begin{figure*}[t]
\centering
\IfFileExists{figures/task_subtask_marginal.pdf}{
  \includegraphics[
    width=\dimexpr\linewidth-2pt\relax,
    trim=1pt 13pt 1pt 1pt,
    clip
  ]{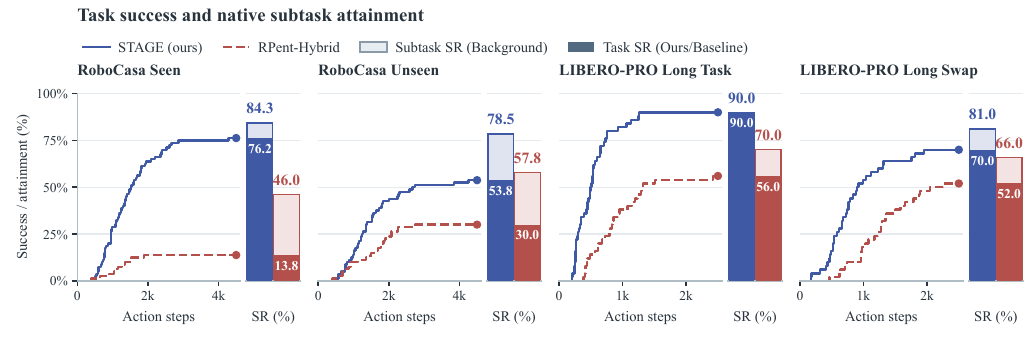}
}{
  \parbox[c][3.4cm][c]{0.94\linewidth}{\centering\small
    \textbf{Figure asset: Task success and native subtask attainment}\par\medskip
    Add the supplied task--subtask figure asset to display the figure.
  }
}
\caption{\textbf{Task success and subtask attainment.}
Curves show cumulative Task SR over action steps. Pale and solid bars show Subtask SR and Task SR, respectively, on the same percentage axis.}
\label{fig:task_subtask_attainment}
\end{figure*}

Figure~\ref{fig:task_subtask_attainment} compares STAGE and RPent-Hybrid across the four main suites using two complementary progress metrics, both evaluated independently of planner-declared completion: the curves trace cumulative Task SR as physical actions are executed, while the background and foreground bars report episode-level Subtask SR (subtask-condition attainment) and final Task SR, respectively. Across all four suites, STAGE achieves most of its task completions early in the rollout and largely plateaus well before the action budget is exhausted, succeeding on 76 task--seed pairs where RPent-Hybrid fails versus failing on only 7 where it succeeds across RoboCasa365 (160 pairs), and 30 vs.\ 4 across LIBERO-PRO (100 pairs).

\paragraph{Task difficulty and recovery frequency.}
Beyond cumulative progress, we analyze how stage-transition behavior relates to task difficulty. At the task level, higher Recovery frequency strongly correlates with lower Task SR: across RoboCasa365 tasks, the Spearman correlation between mean Recovery stages per episode and Task SR is $\rho=-0.86$, and across 20 LIBERO-PRO tasks it is $\rho=-0.70$. Recovery frequency therefore serves as an indicator of task difficulty in the evaluated suites.

Inspecting the recovery-stage traces across the ten low-success RoboCasa365 tasks (Task SR below 50\%) reveals a high intervention load: mean Recovery stages range from 5.8 to 9.6 per episode, and total stages (Advance plus Recovery) range from 7.6 to 13.4 per episode. Code appears in 4.0 to 7.4 Recovery stages per episode, and 2.0 to 5.8 Recovery stages invoke both VLA and Code, with Code accounting for roughly 24\% to 61\% of physical actions inside Recovery stages. While the reused Code-as-Policy interface provides straight-line motion, rotation, and gripper primitives, sliding contact and reliable grasping remain challenging low-level conditions.

\begin{figure}[t]
\centering
\includegraphics[width=\columnwidth]{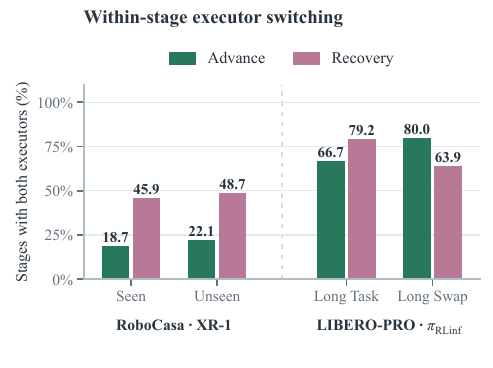}
\caption{\textbf{Within-stage executor switching.} Percentages show nonempty stages whose execution trace contains both VLA and Code, separated by Advance and Recovery transition roles.}
\label{fig:stage_executor_switching}
\end{figure}

\paragraph{Within-stage executor switching.}
Figure~\ref{fig:stage_executor_switching} reports the fraction of nonempty Advance and Recovery stages whose execution trace contains both executors, computed separately per transition role over all successful and failed episodes. On RoboCasa365, mixed execution is more common during Recovery than Advance for both Seen and Unseen suites. On LIBERO-PRO, Recovery switching is higher on Long Task, whereas Advance switching is higher on Long Swap. Across the 80 RoboCasa365 Composite Unseen episodes with XR-1, 71.0\% (480/676) of all physical VLA--Code switches occur within an active stage rather than across Recovery (20.3\%, 137/676) or Advance (8.7\%, 59/676) stage boundaries, confirming that executor reassignment predominantly operates within ongoing stage objectives.

\subsection{Low-Success Tasks and Recovery Frequency}

\begin{figure}[t]
\centering
\setlength{\fboxrule}{1pt}
\setlength{\fboxsep}{0pt}
\includegraphics[width=0.96\columnwidth,trim=5pt 17pt 2pt 5pt, clip]
{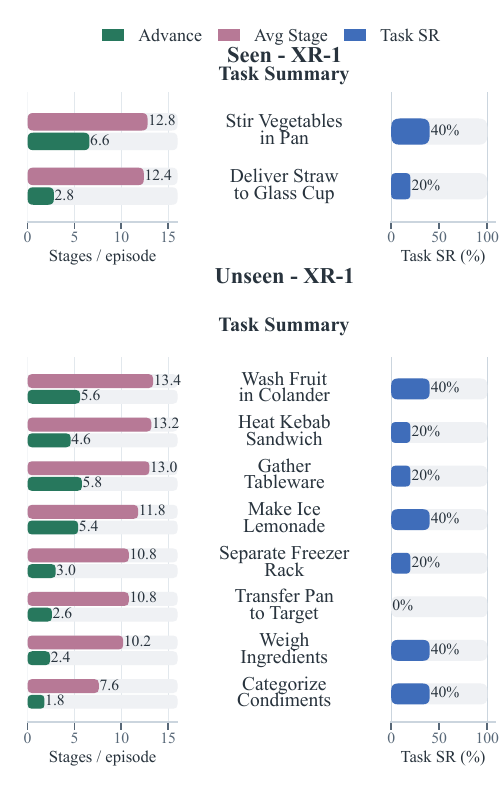}
\caption{\textbf{Low-success RoboCasa tasks.} Green bars show mean Advance stages per episode, purple bars show total stages (Advance plus Recovery), and blue bars show native Task SR. The figure covers every Composite Seen and Composite Unseen task with Task SR below 50\%.}
\label{fig:task_recovery_success_robocasa}
\end{figure}

\begin{figure}[t]
\centering
\includegraphics[width=0.96\columnwidth,trim=5pt 2pt 2pt 5pt, clip]{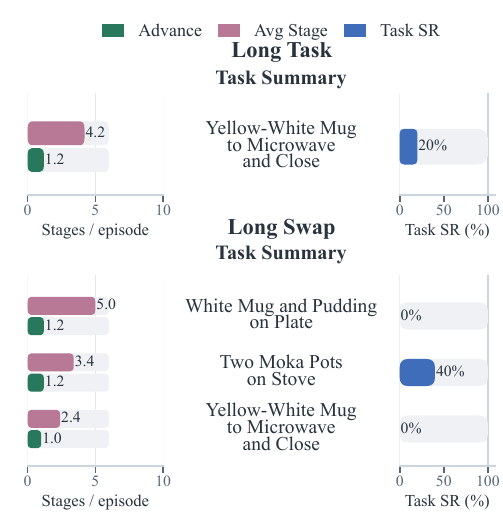}
\caption{\textbf{Low-success LIBERO-PRO tasks.} Follows the same visual encoding as Fig.~\ref{fig:task_recovery_success_robocasa}. The figure covers every Long Task and Long Swap task with Task SR below 50\%.}
\label{fig:task_recovery_success_libero}
\end{figure}

Figures~\ref{fig:task_recovery_success_robocasa} and~\ref{fig:task_recovery_success_libero} display all tasks with Task SR below 50\% (using the same five episodes per task as the main summaries), pairing mean Advance and total stage counts (Advance plus Recovery) on the left with native Task SR on the right.

Across both benchmarks, low-success tasks concentrate in two recurring failure regimes. First, tasks requiring fine contact-rich manipulation or tight-clearance grasping---such as manipulating thin or small items (\textit{Deliver Straw}), sustained sliding or stirring contact (\textit{Stir Vegetables}, \textit{Pan Transfer}), or articulated fixtures (\textit{Separate Freezer Rack}, microwave door closure)---frequently stall at the low-level executor, triggering repeated Recovery stages (5.8--9.6 per episode on RoboCasa365) that geometric primitives alone cannot easily resolve. Second, tasks with visually similar distractors or tightly coupled multi-object placement---such as distinguishing between white and yellow-white mugs under initial-position swaps in LIBERO-PRO Long Swap or coordinating multiple items on a constrained target (\textit{Two Moka Pots on Stove}, \textit{Gather Tableware})---are sensitive to mis-targeting and compounding placement errors. Together with the task-level correlations above, these profiles indicate that remaining failures are driven primarily by low-level contact and grasping bottlenecks and visual disambiguation rather than a lack of planner intervention.

\section{Detailed Ablation and Handoff Analysis}
\label{sec:supp_design_ablations}

The two controlled ablations use the same 16 Composite Unseen tasks and five seeds (80 episodes per variant), keeping XR-1, GPT-6 Astra (low), the 40-action observation interval, scene identities, and physical-action budgets identical to full STAGE.

\paragraph{Global--local coordination.}
At the start of a subtask, the planner selects VLA or Code with no default executor preference. The execution loop locks that choice within the stage, and the prompt maintains the chosen executor until the subtask completes, after which the next subtask permits a new choice; fixed continued instructions and continuous observation history remain unchanged. Preventing within-objective executor handoffs reduces Task SR from 53.75\% to 36.25\% (succeeding on 7 task--seed pairs where STAGE fails vs.\ failing on 21 where STAGE succeeds), while increasing mean stages from 10.05 to 13.59, recovery transitions from 5.70 to 9.78, and physical actions from 1,929 to 1,973 per episode. Without fine-grained handoffs, local stalls trigger repeated recovery cycles rather than in-place executor collaboration.

\paragraph{Selective context retention.}
This ablation reverts continued VLA calls from stage-scoped continuation to stateless per-call invocations. At each continued VLA call, the planner explicitly supplies the instruction and may retain or reformulate it, while the cross-call observation history (images and robot state) is cleared and reinitialized from the current observation rather than carried across calls within the stage. Stage identity, completion criterion, executor switching, and within-call chunk execution (with normal history accumulation inside a single call) remain unchanged. Because policy adapters condition observation history on the active instruction---and invalidate cross-call history when instructions are reissued, as in RPent-Hybrid---removing stage-level call continuity naturally couples instruction persistence with cross-call observation retention. This variant invokes Code in more than 95\% of episodes (78/80), reducing Task SR from 53.75\% to 35.00\% (succeeding on 7 pairs where STAGE fails vs.\ failing on 22 where STAGE succeeds) while increasing mean physical actions from 1,929 to 2,358 per episode.

\paragraph{Instruction continuity across code interventions.}
To measure instruction continuity across VLA--Code--VLA handoffs, we identify every contiguous Code segment immediately preceded and followed by physical VLA execution (perception-only queries, which execute no physical actions, do not split a segment) and compare the two recorded VLA instructions after whitespace normalization. This analysis covers both successful and failed trajectories across all four evaluated suites, measuring textual instruction continuity independently of observation-history retention (since identical wording can also recur across distinct stages). Table~\ref{tab:instruction_continuity} reports the resulting rates.

\begin{table}[t]
\centering
\small
\setlength{\tabcolsep}{3pt}
\caption{\textbf{Instruction continuity at Code handoffs.} Percentage (\%) of VLA--Code--VLA handoffs whose preceding and following VLA instructions match after whitespace normalization.}
\label{tab:instruction_continuity}
\begin{tabular}{@{}lcc@{}}
\toprule
Environment & RPent-Hybrid (\%) & STAGE (\%) \\
\midrule
\begin{tabular}[c]{@{}l@{}}RoboCasa365\\(Seen / Unseen)\end{tabular} & 0.0 / 0.8 & \textbf{36.2} / \textbf{34.1} \\
\addlinespace[3pt]
\begin{tabular}[c]{@{}l@{}}LIBERO-PRO\\(Long Task / Long Swap)\end{tabular} & 16.4 / 16.3 & \textbf{60.7} / \textbf{60.8} \\
\bottomrule
\end{tabular}
\end{table}

Across the matched RoboCasa365 handoffs, STAGE preserves the active instruction in 36.2\% of Seen and 34.1\% of Unseen handoffs, compared with 0.0\% and 0.8\% for RPent-Hybrid. The corresponding LIBERO-PRO rates are 60.7\% and 60.8\% for STAGE, versus 16.4\% and 16.3\% for RPent-Hybrid.

\paragraph{Observed policy-context handling.}
Across the matched XR-1 evaluation subsets, same-stage resumptions after Code include observations captured during code motion. By contrast, RPent-Hybrid's policy interface reinitializes the next VLA call from the current observation after code motion, retaining history only across direct consecutive VLA calls with the same instruction.

On LIBERO-PRO, recorded execution traces confirm that RTC action-context reuse occurs at all same-stage VLA-to-VLA transitions in both Long Task and Long Swap. When Code changes the physical state, the unexecuted portion of the preceding action chunk (which becomes outdated after code motion) is discarded before the next VLA prediction, while same-stage resumption retains the stage instruction. Across backbones, frozen policies expose distinct temporal channels---an observation buffer without RTC on XR-1 and RLDX-1 versus an RTC action tail without an observation buffer on $\pi_{\text{RLinf}}$. Selective Context Retention therefore pairs stage-level instruction persistence with backend-compatible temporal continuity, preserving measured observation history across Code on RoboCasa365 and reusing uninterrupted RTC action tails across same-stage VLA calls on LIBERO-PRO while discarding stale predictions after Code modifies the scene.

\section{Review-Interval Trade-off Analysis}
\label{sec:appendix_interval}

Complementing the normalized retention and cost trends in Figure~\ref{fig:interval} of the main paper, Table~\ref{tab:appendix_interval} reports the underlying absolute success rates and per-task execution costs on RoboCasa365 Composite Unseen. We vary the planner review interval from 40 to 80 and 120 physical action steps while keeping the remaining settings fixed. Subtask SR measures the episode-mean fraction of native conditions attained during execution, as defined in Appendix~\ref{sec:process_annotation}.

\begin{table}[!htbp]
\centering\small
\setlength{\tabcolsep}{4pt}
\caption{\textbf{Observation interval, success, and execution cost.} Cost metrics are averaged per task over both successful and failed episodes. Time cost includes service waiting and tool execution.}
\label{tab:appendix_interval}
\begin{tabular}{@{}lccc@{}}
\toprule
& \multicolumn{3}{c}{Observation interval (steps)} \\
\cmidrule(l){2-4}
Metric & 40 & 80 & 120 \\
\midrule
Task SR (\%) & \textbf{53.75} & 51.25 & 47.50 \\
Subtask SR (\%) & \textbf{78.50} & 77.29 & 74.00 \\
\midrule
Planner requests / task & 72.74 & 48.84 & 43.25 \\
Output tokens / task & 13,758 & 9,533 & 8,449 \\
Time cost / task (min) & 30.3 & 15.3 & 13.6 \\
\bottomrule
\end{tabular}
\end{table}

\section{Implementation Details of STAGE}
\label{sec:supp_impl}

This section complements Section~3 of the main paper by detailing how STAGE decouples a persistent stage contract from local executor reassignment and manages backend-specific VLA context lifecycles alongside the shared observation interval and planner observation window.

\subsection{Stage State and Planner Interface}

\begin{figure*}[!b]
\centering
\includegraphics[width=\linewidth, trim={0pt 14pt 0pt 6pt}, clip]{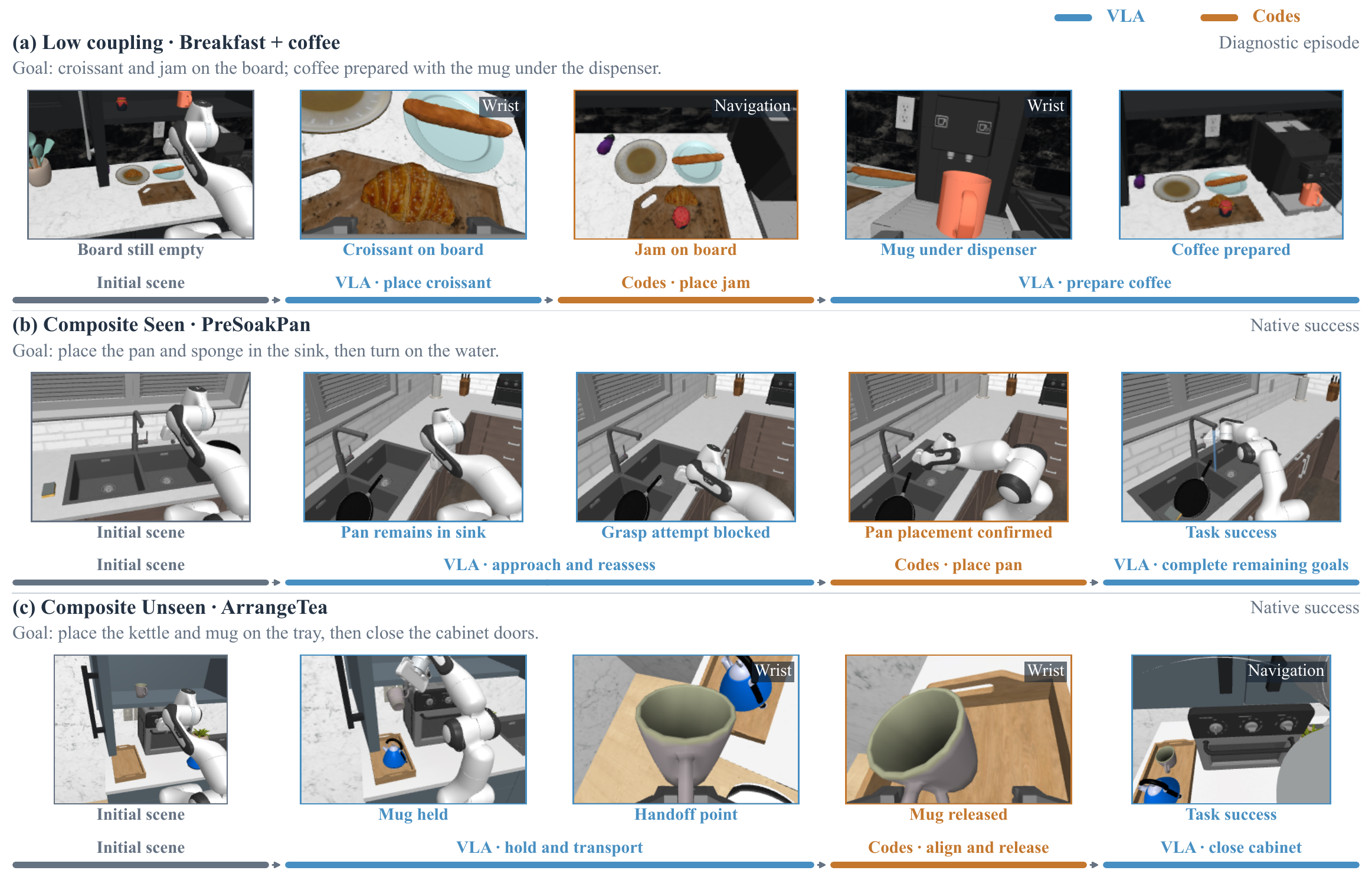}
\caption{\textbf{RoboCasa365 case studies.} Three successful traces covering a low-coupling diagnostic composition, Composite Seen, and Composite Unseen. Each trace includes both VLA and Code execution.}
\label{fig:supp_robocasa_cases}
\end{figure*}

Each stage stores an instruction, a completion criterion, and a transition history. An initial stage is instantiated at the start of each episode. \textsc{Continue} retains the active instruction and criterion, while \textsc{Advance} and \textsc{Recover} create a new stage and initialize its policy context. A recovery also records the preceding stage as its parent.

A review records the planner's assessment of progress, visibility, observation evidence, and transition evidence. Executor selection is expressed through subsequent motion calls or generated programs, separately from the stage-transition label. A \textsc{Continue} decision can therefore be followed by VLA or Code execution, and executor handoffs preserve the active stage and its action allowance.

Stage consistency is enforced structurally: a continued stage retains its instruction and completion criterion, and VLA calls use that accepted instruction. Planner assessments remain recorded judgments, whereas evaluation uses native benchmark outcomes and the separately defined progress metrics.

\subsection{Shared Observation Interval}

The execution loop is given in Algorithm~\ref{alg:stage} of the main paper. This section specifies how its shared action allowance is enforced.

A single action counter bounds motion by both executors between planner observation reviews. The limit is checked at every environment step, including VLA chunks, individual analytic primitives, and sequences of calls in a generated program. On exhaustion, execution returns for observation and review before more motion is allowed. A fresh reviewed observation resets the allowance, while repeated calls on the same observation do not create additional physical-action budget. Subsequent analytic motion is selected from the updated observation.

\subsection{VLA Context by Backend}

STAGE retains or reinitializes whichever temporal context channel is exposed by the frozen policy's official inference interface---such as an observation history buffer or RTC action-chunk conditioning.

\paragraph{Observation-history context.}

\begin{figure*}[t]
\centering
\includegraphics[width=\linewidth, trim={0pt 14pt 0pt 6pt}, clip]{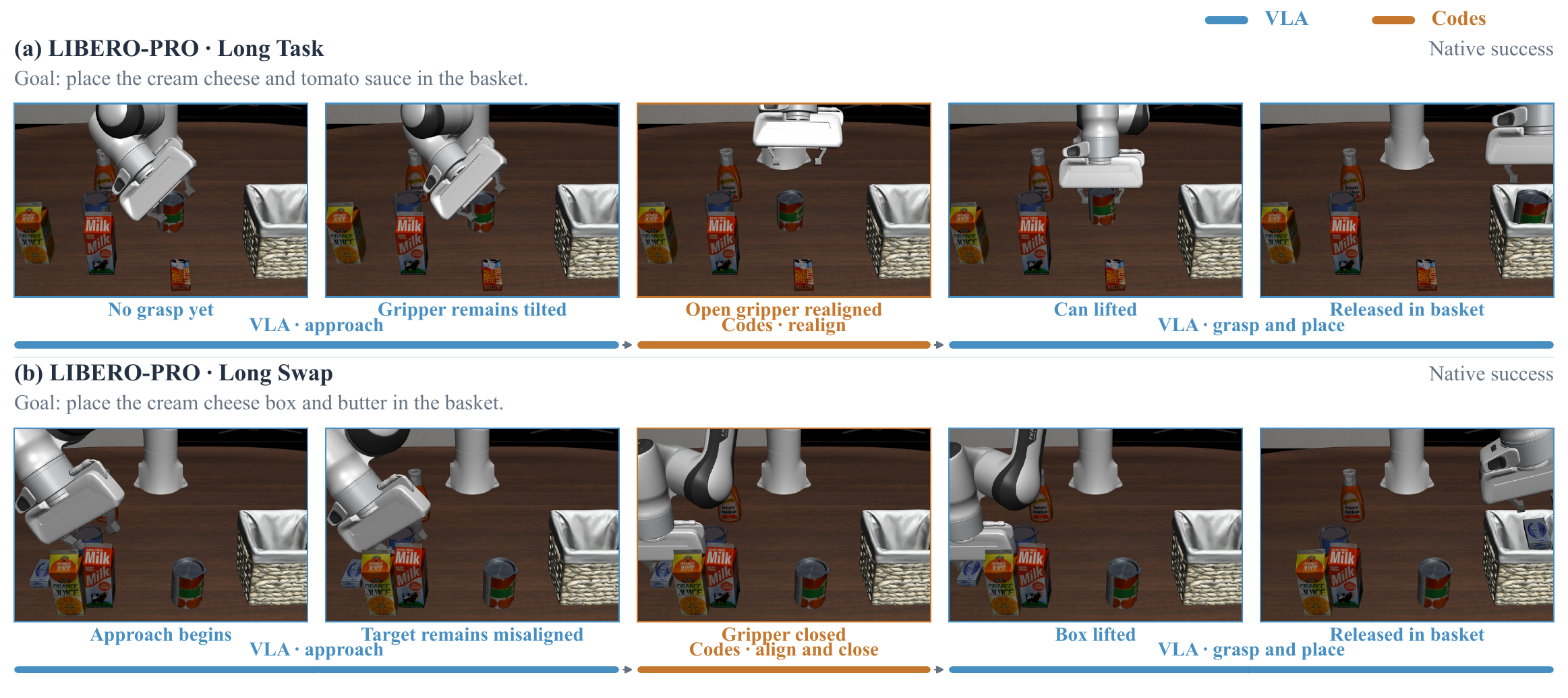}
\caption{\textbf{LIBERO-PRO case studies.} Successful Long Task and Long Swap traces with VLA--Code handoffs.}
\label{fig:supp_libero_cases}
\end{figure*}

Both RoboCasa365 policy interfaces condition on a finite temporal image buffer. XR-1 also incorporates robot-state history, whereas RLDX-1 uses only the current robot state. At a new stage, the buffer is cleared and initialized from the current observation, and subsequent measured observations replace the initialization. Within a continued stage, repeated VLA calls use the evolving observation buffer, including observations recorded during intervening analytic actions. If Code executes before a stage's first VLA call, that call initializes from the latest post-code observation.

\paragraph{Action-chunk context.}
The LIBERO-PRO policy conditions on current images and robot state, with temporal continuity provided by Real-Time Chunking (RTC)~\cite{black2025rtc}, which conditions each prediction on the unexecuted portion of the preceding action chunk (prediction horizon of ten, execution prefix of five, synchronous zero-delay execution). Continued VLA calls within the same stage reuse this action context. When Code modifies the physical state, the outdated unexecuted action tail is discarded before the next VLA prediction while keeping the stage instruction intact; a new stage likewise starts without the preceding stage's tail.

\paragraph{Policy-context reinitialization.}
Reinitializing policy context updates local policy inputs while leaving simulator state and model parameters unchanged. For observation-history policies, reinitialization starts from the current observation until measured history accumulates; for RTC policies, it predicts from current inputs without a prior action tail. These backend-specific updates can occur within a stage whenever physical state changes invalidate prior predictions, while the stage objective remains fixed.

\subsection{Planner Observation Context}

The VLM receives the current stage instruction and criterion together with visual execution evidence. The default image context contains the stage's initial observation and up to four recent observation timestamps, with duplicate timestamps removed. A new stage replaces the reference observation and starts a new image window. Textual conversation and tool records remain available, while older images are replaced with text placeholders in the delivered history. This planner image window is separate from the VLA sensor buffer and RTC action context.

The RoboCasa365 planner uses external, navigation, and wrist views. The LIBERO-PRO interface uses external and wrist views, plus an additional external camera view used by the policy. The reference observation is captured at the start of the current stage and supports within-episode visual assessment.

\section{Case Studies}
\label{sec:supp_coupled_protocol}

The main analysis uses the high-coupling board--waffle trace to illustrate how stage-based coordination handles a shared-object dependency. This appendix adds successful recorded cases and separates the RoboCasa365 and LIBERO-PRO execution patterns. These qualitative examples illustrate representative coordination behaviors rather than establishing causal attribution on their own, and do not change the aggregate evaluation metrics.

\subsection{RoboCasa365}

Figure~\ref{fig:supp_robocasa_cases} shows three successful RoboCasa365 traces: the low-coupling breakfast--coffee diagnostic case, along with benchmark-valid Composite Seen and Composite Unseen episodes. Each trace includes both VLA and Code execution. In the low-coupling case, VLA places the croissant, Code places the jam, and VLA prepares the coffee. In Composite Seen, VLA approaches the pan, Code completes the pan placement, and VLA finishes the remaining sink and water conditions. In Composite Unseen, VLA carries the mug toward the tray, Code completes the release, and VLA closes the cabinet.

The RoboCasa365 traces expose kitchen-specific handoffs around object placement and navigation. Task success for every displayed episode is determined by the benchmark's native evaluator. The low-coupling diagnostic case is excluded from the aggregate benchmark metrics, whereas the Composite Seen and Composite Unseen examples are drawn directly from the evaluated test suites.

\subsection{LIBERO-PRO}

Figure~\ref{fig:supp_libero_cases} shows one successful Long Task trace and one successful Long Swap trace. Both traces include VLA and Code execution within the same task objective. In Long Task, Code realigns the open gripper and VLA resumes the placement. In Long Swap, Code corrects the approach and closes the gripper before VLA completes the basket placement while the earlier condition remains satisfied.

The LIBERO-PRO traces make the simulator-specific behavior explicit: a Code pose correction is followed by VLA continuation on the fixed tabletop scene, with the RTC policy context updated at the next VLA call. Together with the RoboCasa365 cases, the examples show how the same stage objective can be continued across executor changes under different observation and control interfaces.

\section{Code and Rollout Availability}
\label{sec:supp_release}

The evaluation code and configuration files used for the reported experiments are publicly available at \url{https://github.com/illimit3/stage-robotics}. Complete rollout trajectories will be released upon publication.

\end{document}